\documentclass[conference]{IEEEtran}

\IEEEoverridecommandlockouts

\usepackage{cite}
\usepackage{amsmath,amssymb,amsfonts}
\usepackage{graphicx}
\usepackage{subfigure}
\usepackage{textcomp}
\usepackage{xcolor}
\usepackage{algorithm}
\usepackage{algorithmicx}  
\usepackage{algpseudocode}  
\usepackage{amsmath}
\usepackage{hyperref}

\newcommand{\fig}[1]{Fig.~\ref{fig:#1}}

\def\BibTeX{{\rm B\kern-.05em{\sc i\kern-.025em b}\kern-.08em
    T\kern-.1667em\lower.7ex\hbox{E}\kern-.125emX}}

\title{Vision Based Picking System for Automatic Express Package Dispatching\\}

\author{Shengfan Wang$^{\dagger}$, Xin Jiang$^{*}$, Jie Zhao, Xiaoman Wang, Weiguo Zhou and Yunhui Liu, Fellow, IEEE%
\thanks{This work was supported by the following projects: Shenzhen Peacock Plan Team grant (KQTD20140630150243062), Shenzhen and Hong Kong Joint Innovation Project (SGLH20161209145252406), Shenzhen Fundamental Research grant (JCYJ20170811155308088).
}%
\thanks{Shengfan Wang, Xin Jiang, Jie Zhao, Xiaoman Wang and Weiguo Zhou are with the School of Mechanical Engineering and Automation, Harbin Institute of Technology, Shenzhen 518055, China. The author e-mail: 18S053234@stu.hit.edu.cn. The corresponding author email: x.jiang@ieee.org.}%
\thanks{Yunhui Liu is with the Department of Mechanical and Automation Engineering, The Chinese University of Hong Kong, Shatin, Hong Kong, China.}%
}

\begin{document}
\maketitle

\begin{abstract}
This paper presents a vision based robotic system to handle the picking problem involved in automatic express package dispatching. By utilizing two \textbf{RealSense} RGB-D cameras and one UR10 industrial robot, package dispatching task which is usually done by human can be completed automatically. In order to determine grasp point for overlapped deformable objects, we improved the sampling algorithm proposed by the group in Berkeley to directly generate grasp candidate from depth images. For the purpose of package recognition, the deep network framework YOLO is integrated.
  We also designed a multi-modal robot hand composed of a two-fingered gripper and a vacuum suction cup to deal with different kinds of packages. All the technologies have been integrated in a work cell which simulates the practical conditions of an express package dispatching scenario. The proposed system is verified by experiments conducted for two typical express items.
\end{abstract}

\section{Introduction}
Recently, in many industry fields, human beings are replaced by robots. Although robot based automation has achieved great progress, there are still many tasks necessitate human labor. In logistics field, Amazon had held the robot competition (ARC) aiming to solve the problems involved in logistic process \cite{correll2018analysis,eppner2016lessons,corbato2018integrating}. When Amazon first held the Amazon Picking Challenge in 2015, many teams failed to pick the specified items by their robotic systems. At that time, many teams only employed two fingered gripper as the end effector. While in the 2017 Amazon Robotics Challenge, most teams achieved high marks and employed both gripper and vacuum suction cup. It was proved in the challenge that the combination of gripper and vacuum suction cup was efficient. Although the challenge is terminated, it attract interests from the researchers in finding robotic solution in  
logistics.

One of the tasks in logistic system which is considered as a potential target for robotic automation is the process of package dispatching. Currently, this process is conducted by human workers. In the process, the worker have to pick up the package from the conveyor and recognize the destination information printed on it. Based on the information, the worker then dispatches the package to the specified line. The main problems involved in the procedure is that the label printed on the package may not face upwards. In this case, the sensors can not capture the label and it needs the robot to grasp the package and then reverse it. In many situations, the packages are crowded randomly. This increases the difficulty in finishing the item grasping. In this research, we try to tackle the problems confronted in this situation by using RGB-D cameras.

\section{Related Work}
For decades, researchers have been doing research on robot grasping \cite{bicchi2000robotic,shimoga1996robot,sahbani2012overview,bohg2014data,caldera2018review}. In the early work \cite{saxena2007robotic,saxena2008robotic} , human-designed features were used to represent grasps in images. In order to generate grasp candidate, full 3-D model of objects is necessary\cite{miller2003automatic}. These methods was popular at that time, but they all faced challenges in the situations where no robust features nor full 3-D models are available. These situations are common in practical applications. Recently, many algorithms are proposed by to solve the robotic grasping  problem without using CAD models. Andreas et al. \cite{gualtieri2016high,Pas2017Grasp} proposed to directly generate grasp poses in point clouds and then use neural networks to rank all candidate poses in order to select the most proper one. Lenz et al. \cite{lenz2015deep} attempted to employ deep learning technologies to learn and detect features representing proper robotic grasps using RGB-D images. In Amazon Robotics Challenge, team MIT-Princeton \cite{zeng2017multi} built a multi-view vision system to estimate the 6D pose of objects. They also studied the policy of utilizing two functioned hand composed of two-fingered gripper and vacuum suction cup \cite{zeng2018robotic}  Causo et al. \cite{inproceedings} and Eppner et al. \cite{eppner2016lessons} designed robust robot systems for item picking integrated with perception, motion planning and special purpose end effector. Wade et al. \cite{wade2017design} designed a multi-modal end-effector which is combined with three grasp synthesis algorithms. 

The group in Berkeley did a lot of researches on robust grasp planning. They build a large dataset for grasp and suction planning which is called Dexterity Network. The dataset is composed of huge sensor data in various kinds of grasping scenarios and the corresponding metrics for evaluating grasp candidates. Both the grasping methods with parallel-jaw gripper and that using vacuum suction cup are considered in the dataset. They proposed the Grasp Quality Convolution Neural Network for evaluating grasp candidate directly from sensor input. This neural network is trained on the dataset of Dex-Net. The grasp and suction evaluation algorithm directly samples candidates from depth image with no object models assumed and this feature makes the method capable of dealing with novel objects. 

Our work is inspired mainly by the algorithms proposed by the group in Berkeley and we have made the following contributions. Firstly, we improve the grasp sampling algorithm and it demonstrated better performance compared with the original one when dealing with the picking problems confronted with express package dispatching application. Secondly, we design a two-functioned robot hand consisting of a two-fingered gripper and a vacuum suction cup. Finally, by combining the methods for object detection YOLO\cite{redmon2016you,redmon2017yolo9000} with the Open Source Robot Operating System, we integrated a robot system shown in \fig{system}, with which a typical express package dispatching demonstration is realized. 

\section{Problem Statement}
\begin{figure}[tb]
\centering 
\subfigure[Express envelope]{\includegraphics[width=4cm,height=2.5cm]{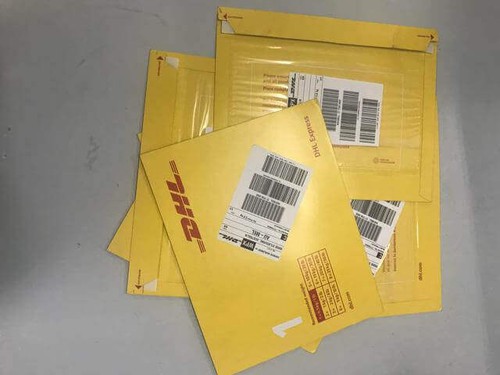}}
\subfigure[Express bag]{\includegraphics[width=4cm,height=2.5cm]{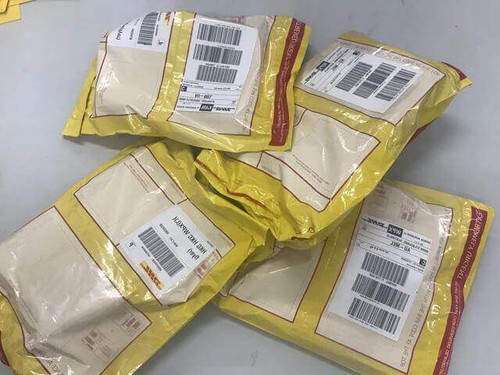}} 
\caption{The express bag and envelope tackled in the dispatching procedure} 
\label{fig:real_status}
\end{figure}

In the subsequent demonstration of package dispatching procedures, two kinds of items: bags and envelopes are considered. In practical dispatching line, they are processed separately.
The respective conveyors transmit items to dispatching work cell. A worker has to recognize the information printed on the label of the package then dispatch them according to the
information recognized. Since the status of the packages on the conveyor are random and overlapped with each other as shown in \fig{real_status}, the workers have to reverse the package if
it does not face upward in order to see the label. The difficulties in a robotized solution for this procedure mainly come from the task of picking one package from overlapped ones. A successful
dispatching requires that for all the packages the barcodes printed on them face upward and are recognized correctly. 

\section{The Approach}
\begin{figure}[tb]
\centering 
\subfigure{\includegraphics[width=7cm,height=6cm]{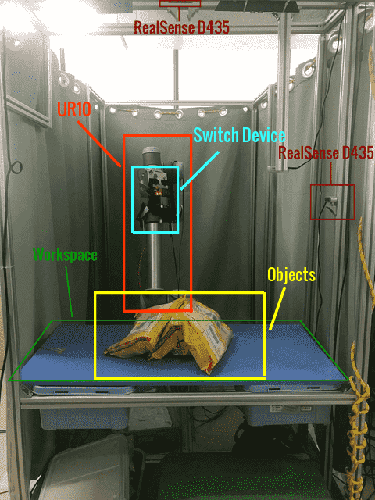}} 
\caption{The robot system} 
\label{fig:system}
\end{figure}

\begin{figure}[tb]
\centering 
\subfigure{\includegraphics[width=7cm,height=6cm]{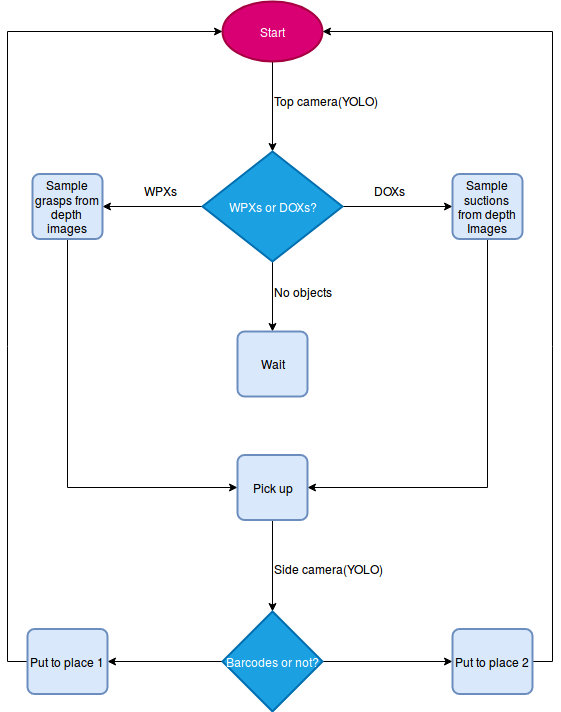}} 
\caption{The flowchart of the processing in the express package dispatching system} 
\label{fig:pipeline}
\end{figure}

Our automatic express dispatching system is composed of two components, the vision processing system and the manipulator. As for the vision processing system, two RealSense D435 cameras are employed to provide color and depth images. The color images are used to detect objects and recognize the barcodes, while the depth images are fed to grasping planning algorithm. As for the manipulator, we use one UR10 robot mounted with an end-effector which support the usage of switching between using two-fingered gripper and a vacuum suction cup. All the programs of the system are implemented with Robot Operating System. \fig{system} shows the whole system and the flowchart of the information processing pipeline is shown in \fig{pipeline}.
\subsection{Package Recognition Method}
The neural network frame YOLO is popular in object detection field for its accuracy and real-time performance when compared to Faster R-CNN\cite{ren2015faster}. For the purpose of package recognition,
we tried both the official implementation of YOLO and Faster R-CNN based detection methods from Darknet and Google. They both perform well for our tasks. But there is a trade-off between the accuracy and speed. Faster R-CNN achieves higher accuracy while YOLO demonstrates faster speed. Considering the requirement of the real time performance in practical applications, we finally choose YOLO as the detector.

For the target of envelope and bag, we prepared for 25 pictures for each category respectively. Each picture contains multiple objects and all the objects are placed randomly like the situations of real industrial environment. We resize each picture to 640 $\times$ 480 and image argumentation is conducted to the dataset by random rotating each picture. It finally results to a dataset consisting of 300 images.

\begin{figure}[tb]
\centering 
\subfigure[envelope]{\includegraphics[width=4cm,height=2.5cm]{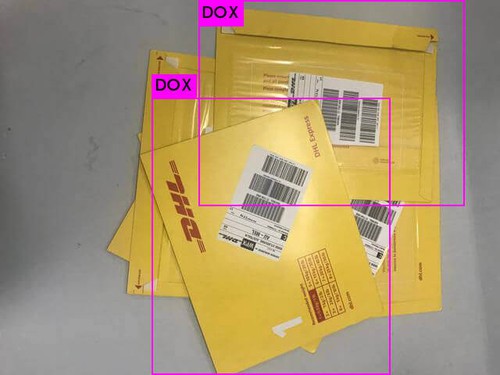}} 
\subfigure[bag]{\includegraphics[width=4cm,height=2.5cm]{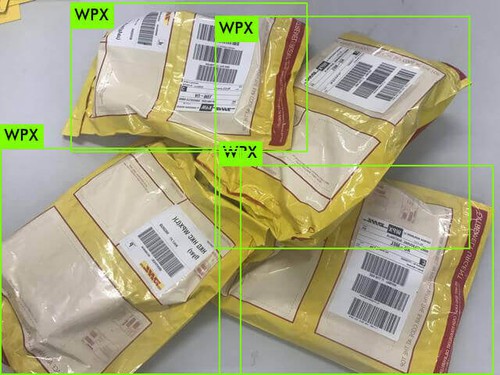}} 
\caption{Envelop and bag recognition} 
\label{fig:bag_reg}
\end{figure}

\begin{figure}[tb]
\centering 
\subfigure{\includegraphics[width=4cm,height=2.5cm]{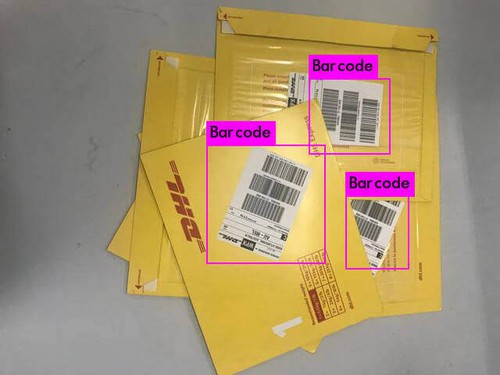}} 
\subfigure{\includegraphics[width=4cm,height=2.5cm]{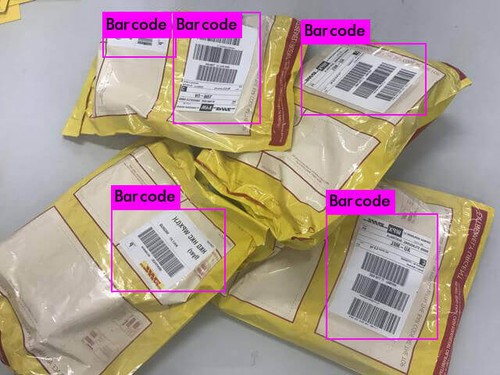}}   
\caption{Barcodes recognition} 
\label{fig:barcode}
\end{figure}

In order to detect express packages and barcodes, two YOLO networks are trained separately. One of them is used for package detection and the other is used for barcode recognition. The processing
pipeline is designed assuming that the robot first picks one item up using the camera configured above the workspace and then the other camera configured next to the workspace will be triggered to detect whether there is a barcode. Based on the recognition result, the robot will decide how to process the object. For those packages which barcodes face upward, the robot will place them on the conveyor directly, otherwise the object will be reversed to make its barcode face upward. After training of 20000 epochs, the networks achieves good performance in distinguishing envelopes from bags as shown in \fig{bag_reg} and barcode recognition shown in \fig{barcode}.
\subsection{Grasp Policy}
\begin{figure}[tb]
	\centering 
	\subfigure{\includegraphics[width=8cm,height=2.5cm]{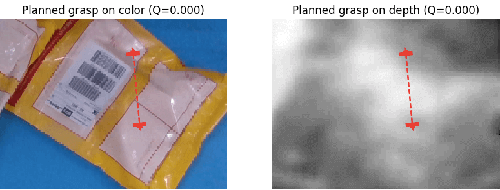}}
	\caption{The execution of the grasp may cause collision between the robot end-effector and the object enclosed inside the bag} 
	\label{fig:test}
\end{figure}

Considering the deformable property of the express bags, we choose to use two-fingered gripper to grasp them. The grasp planning is implemented following the method which samples antipodal grasps
directly from depth images\cite{mahler2017dex}. After obtaining hundreds of candidate grasps, the GQ-CNN\cite{mahler2017dex} is used to rank them and choose the best one.
When we use the original grasp policy proposed by Berkeley, we find that it is likely to generate unreasonable results, which will lead to failure of grasping. In addition, we do not want the
sampled grasp candidates to be on the surface of the bags, since it may lead to the collision between end-effector and items enclosed inside the bag, as shown in \fig{test}. Without information
on the objects inside the bag, the grasp plan indicates danger. Thus we attempted to improve the algorithm by taking all above considerations into account, and its detail is shown in Algorithm \ref{algorithm}.

\floatname{algorithm}{Algorithm}  
\renewcommand{\algorithmicrequire}{\textbf{Input:}}  
\renewcommand{\algorithmicensure}{\textbf{Output:}}  

\begin{algorithm}
\caption{Improved grasp sampling algorithm}
\begin{algorithmic}[1]
  \Require The set of candidate grasps generated by the original antipodal grasp sampling algorithm is denoted as $\mathcal{G}$. The depth value at the two parallel jaws position are represented separately as $d_1$ and $d_2$.
  The depth value at the grasp center is denoted as $d_0$. The color value at the two parallel jaws position are denoted separately, as $c_1$ and $c_2$. The mean and standard deviation value of depth and color
  in the region covered by the grasp are denoted as $ \mu_{d}, \mu_{c}, \delta_{d}, \delta_{c}$. The maximum and minimum depth value in the region covered by the grasp, are denoted as $d_{\max}$ and $d_{\min}$ .  The parameters, $ \epsilon$ in the following represents threshold values.
\Ensure The final grasp candidates set, $\tilde{\mathcal{G}}$, 
\For{every $g \in \mathcal{G}$}
	\If{$d_1 >  d_0 + \epsilon_1$ and $d_2 >  d_0 + \epsilon_1$}	
		\If{$ d_{\max} - d_{\min}  > \epsilon_2$}	
			\If{$\mu_{d} > d_0 + 	\epsilon_3$ and $\delta_d > \epsilon_4$}			
				\If{$\delta_{c} > \epsilon_5$}			
					\If{mean($c_1 - c_2)> \epsilon_6$}				 
						\State $g$  $\cup$ $ \tilde{\mathcal{G}} $					
					\EndIf				
				\EndIf			
			\EndIf		
		\EndIf	
	\EndIf
\EndFor
\State return $\tilde{\mathcal{G}}$
\end{algorithmic}
\label{algorithm}
\end{algorithm}
The improved algorithm works like a filter and it ensures that only the reasonable grasp candidate will be left. The method is inspired by \cite{gualtieri2016high,Pas2017Grasp}. The idea behind the constrains described in the algorithm is to make sure that there is one part of the object in the grasp region as well as the color and depth distribution are different between the candidate grasp region and that of the outside. To be more specific, the depth and color value at the parallel jaw positions, grasp center positions are different with that outside of the parallel jaws. The optimal grasp for a bag with no information about the object enclosed in it is to grasp its corner. The proposed algorithm is to filter out the candidate satisfying the requirement.   
		
\begin{figure}[tb] 
\centering 
\subfigure[The first grasp]
{\includegraphics[width=4cm,height=3cm]{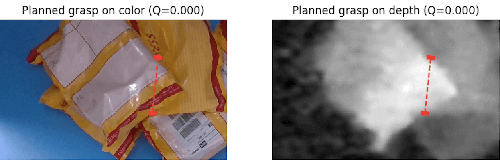} } 
\subfigure[The second grasp]{\includegraphics[width=4cm,height=3cm]{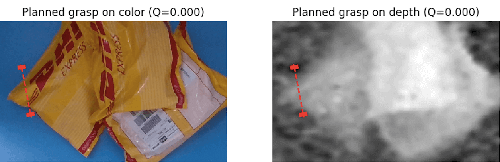} } 
\subfigure[The third grasp]
{\includegraphics[width=4cm,height=3cm]{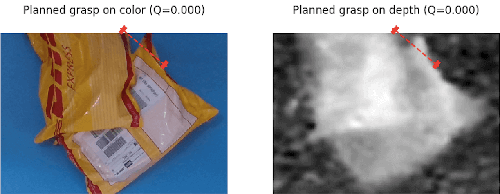} }
\subfigure[The fourth grasp]
{\includegraphics[width=4cm, height=3cm]{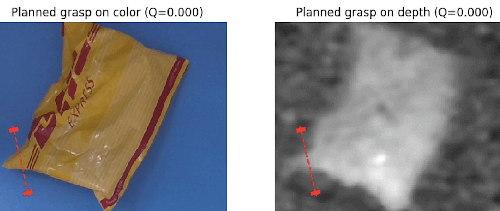} }
\caption{The generated grasp plan during a whole subsequent picking process of four overlapped bags.} 
\label{fig:wpx_demo}
\end{figure}

\begin{figure}
\centering 
\subfigure{\includegraphics[width=4cm,height=2.5cm]{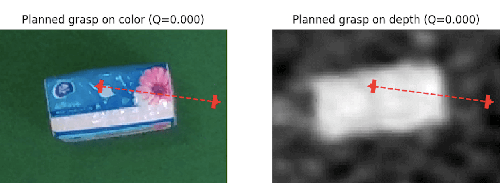} }  
\subfigure{\includegraphics[width=4cm,height=2.5cm]{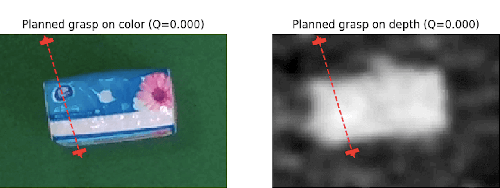} } 
\subfigure{\includegraphics[width=4cm,height=2.5cm]{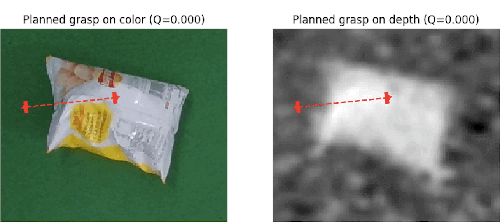} }
\subfigure{\includegraphics[width=4cm,height=2.5cm]{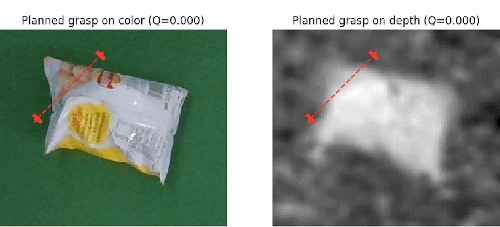} }
\subfigure{\includegraphics[width=4cm,height=2.5cm]{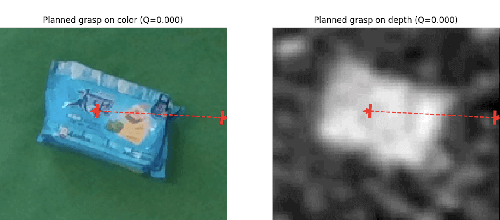} }
\subfigure{\includegraphics[width=4cm,height=2.5cm]{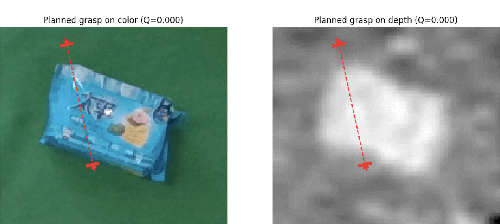} }
\subfigure{\includegraphics[width=4cm,height=2.5cm]{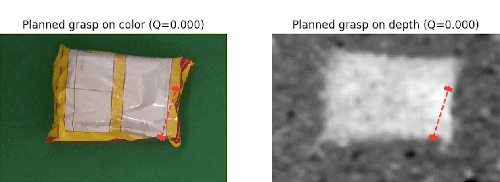} }
\subfigure{\includegraphics[width=4cm,height=2.6cm]{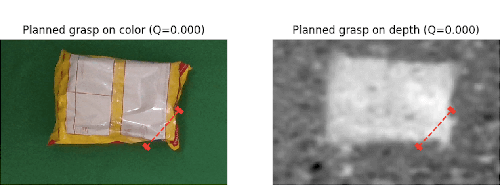} }
\caption{Comparison between the original sampling algorithm(left) and the improved one (right)} 
\label{fig:comparison}
\end{figure}

\fig{wpx_demo} illustrates the whole sampling results from the improved algorithm. We could see that the algorithm outputs grasp candidates around the four corners of the bag. They are safer with low possibility of leading to collision between the robot hand and the objects inside the bag. The comparison between the proposed algorithm and the original one is demonstrated in \fig{comparison}. It is demonstrated that the proposed algorithm generate more reasonable grasp plan than that generated with the original one. We argue that the proposed method can also be applied to other similar situations.

\subsection{Suction Policy}
The envelopes are flat, which makes it suitable for using suction cup to directly suction on the surface in order to pick them up. We utilize the suction sampling policy proposed in the work \cite{mahler2017suction}. We also assumes that the position near the center of the envelope is the ideal place for suction. Therefore, we use YOLO as the detector to find where the envelope is and then takes sampling suction point near center of the detection result. We use a vacuum cleaner to provide the necessary suction force. This kind of utilization is also popular in Amazon Picking Challenge \cite{inproceedings,morrison2018cartman,
zeng2017multi,correll2018analysis,eppner2016lessons,wade2017design,yu2016summary,corbato2018integrating}.

\subsection{Gripper and Suction Cup Switch Device}
\begin{figure}[tb]
\centering 
\subfigure[suction model]{\includegraphics[width=4cm,height=4.5cm]{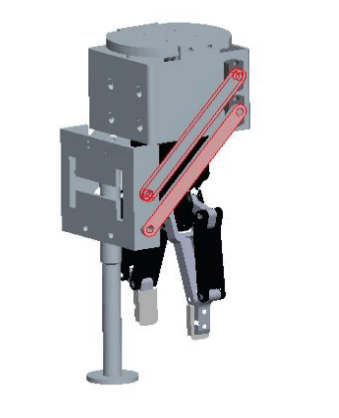}} 
\subfigure[grasp model]{\includegraphics[width=4cm,height=4.5cm]{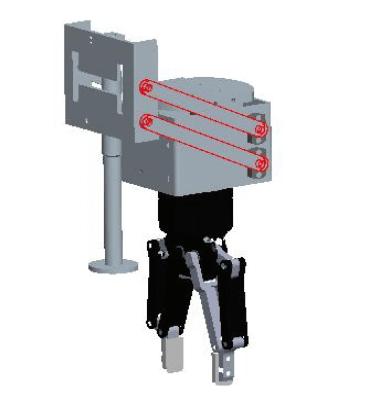} } 
\caption{Gripper and suction cup in two operation modes} 
\label{fig:switch_device_two}
\end{figure}

In Amazon Picking Challenge, many teams employ a device to switch between gripping and suction mechanism in order to choose different picking ways according to the property of targets\cite{correll2018analysis,wade2017design,yu2016summary,corbato2018integrating,morrison2018cartman}. We also designed such a device to enable switch between
a Robotiq 140 two-fingered gripper and Schmalz's suction as shown in \fig{switch_device_two}.

Different from the other similar devices, our switch device also supports the usage of simultaneously employing both of the two mechanisms. In some situations, this provides merit for object manipulation. For example, for picking a book placed on a table, we can choose to firstly suction it. Since generally the width of the book will be beyond the that of the griper jaws, it is impossible for gripping. After the book suctioned, we can then make the gripper to achieve a stable grasping.  

\section{Experiment}
\begin{figure}[tb]
\centering 
\subfigure{\includegraphics[width=4cm,height=3cm]{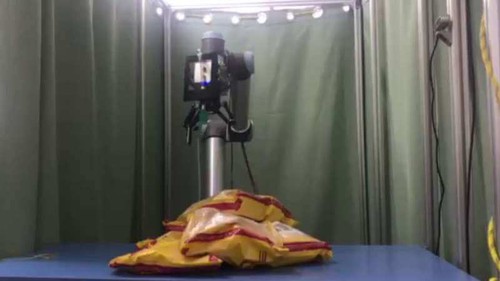}} 
\subfigure{\includegraphics[width=4cm,height=3cm]{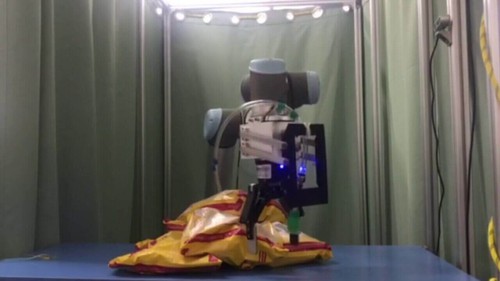}} 
\subfigure{\includegraphics[width=4cm,height=3cm]{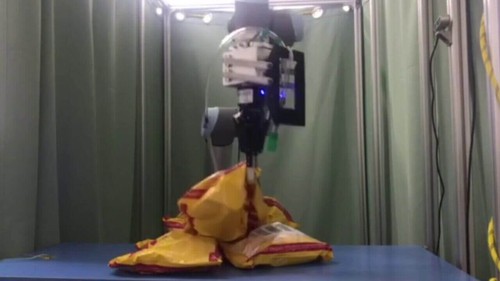}} 
\subfigure{\includegraphics[width=4cm,height=3cm]{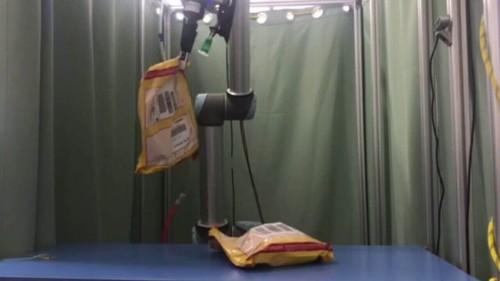}} 
\subfigure{\includegraphics[width=4cm,height=3cm]{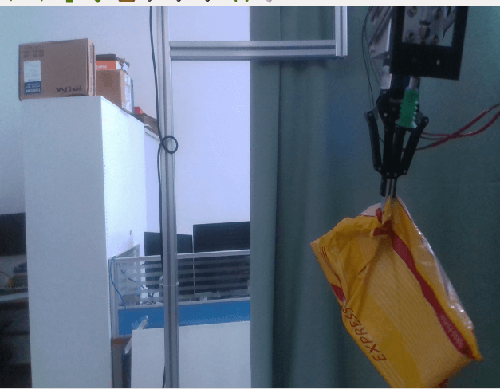}} 
\subfigure{\includegraphics[width=4cm,height=3cm]{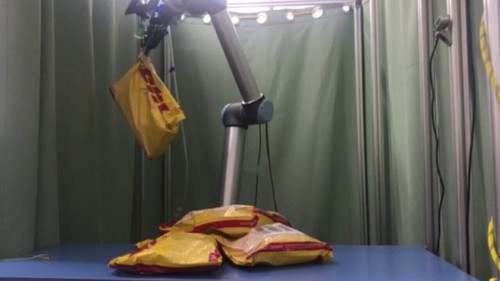}} 
\caption{Express bags picking experiment} 
\label{fig:wpx_system}
\end{figure}

\begin{figure}[tb]
\centering 
\subfigure{\includegraphics[width=4cm,height=3cm]{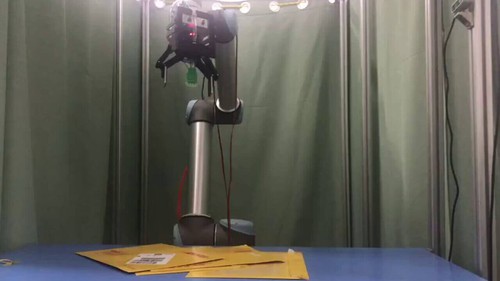}} 
\subfigure{\includegraphics[width=4cm,height=3cm]{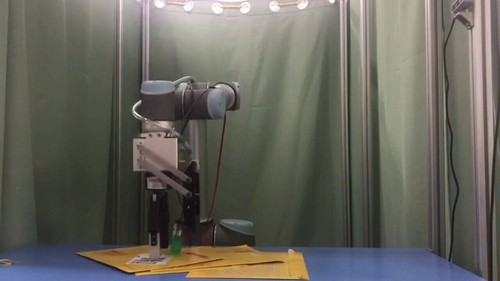}} 
\subfigure{\includegraphics[width=4cm,height=3cm]{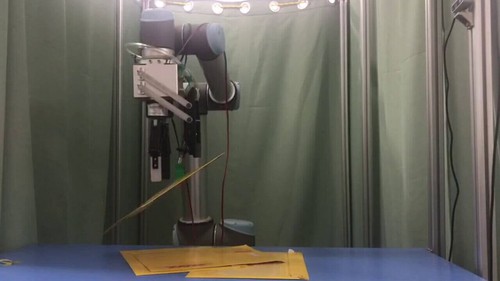}} 
\subfigure{\includegraphics[width=4cm,height=3cm]{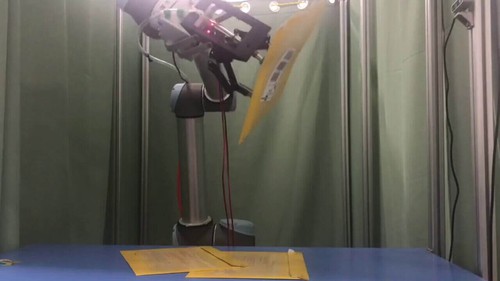}} 
\subfigure{\includegraphics[width=4cm,height=3cm]{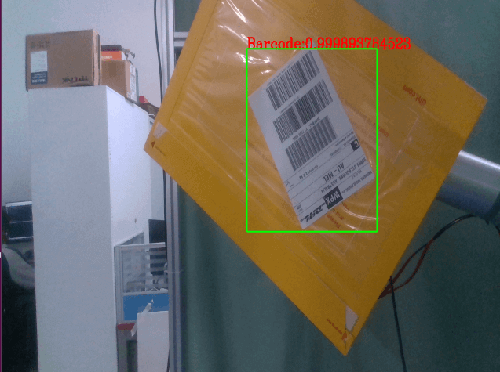}} 
\subfigure{\includegraphics[width=4cm,height=3cm]{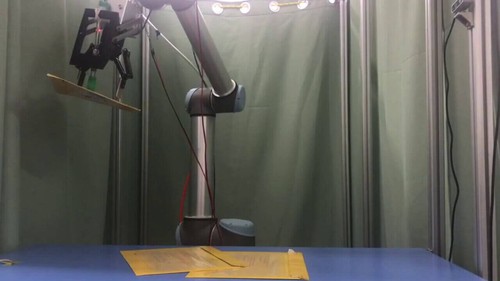}} 
\caption{Express envelopes picking experiment} 
\label{fig:dox_system}
\end{figure}

We implement all the proposed methods under ROS and conducted the corresponding experiments with our platform with a single GPU (NVIDIA GeForce GTX 1060). The experiment video can be viewed here \footnote{\url{https://youtu.be/TgD2G8B-QSY}}. In the verification experiments, we set the parameters in algorithm \ref{algorithm} as follows: $\epsilon_1 = 0.01$, $\epsilon_2 = 0.01$, $\epsilon_3 = 0.01$, $\epsilon_4 = 0.01$, $\epsilon_5 = 30$, $\epsilon_6 = 50$. \fig{wpx_system} and \fig{dox_system} show the whole grasping and recognition procedure for express envelope and bag respectively. In each of the experiments, four objects are placed on the table in the initial state. The robot then conducted dispatching procedure of "pick-recognize-place" one by one. For the envelopes, the robot would choose to use suction and its success rate is almost 100\%. For the bags, the robot would choose two-fingered gripper to complete the task. Its success rate is about 90\%. The failures in the experiments are mainly due to the slip between the object and the gripper which leads to object dropping in transmitting phase. The time cost of picking and placing four bags is about fifty seconds and is about one minute and a half for four envelopes. For further study, we will try to reduce the time cost of whole process.

\section{Conclusion}
In this research, we proposed a robot system to tackle the problem in express packages dispatching. In this procedure, a human worker has to pick up the packages and recognize the label on the packages. Then he should dispatch them according to the recognized information. The difficulty involved in the procedure is that the packages are randomly placed and overlapped with each other. In addition, without prior information on the objects enclosed inside the package, grasping may fail due to the contact between the robot hand and the object inside. For this problem an improved grasping planning method is proposed which generates grasp plan avoiding direct contact to the center of the package. For the purpose of package recognition, neural network based detection method is integrated. With the proposed methods, the robot system demonstrated in experiments the capability of picking and recognizing two typical express packages: envelope and bag.

\bibliography{ref1}

\begin{thebibliography}{10}
\providecommand{\url}[1]{#1}
\csname url@samestyle\endcsname
\providecommand{\newblock}{\relax}
\providecommand{\bibinfo}[2]{#2}
\providecommand{\BIBentrySTDinterwordspacing}{\spaceskip=0pt\relax}
\providecommand{\BIBentryALTinterwordstretchfactor}{4}
\providecommand{\BIBentryALTinterwordspacing}{\spaceskip=\fontdimen2\font plus
\BIBentryALTinterwordstretchfactor\fontdimen3\font minus
  \fontdimen4\font\relax}
\providecommand{\BIBforeignlanguage}[2]{{%
\expandafter\ifx\csname l@#1\endcsname\relax
\typeout{** WARNING: IEEEtran.bst: No hyphenation pattern has been}%
\typeout{** loaded for the language `#1'. Using the pattern for}%
\typeout{** the default language instead.}%
\else
\language=\csname l@#1\endcsname
\fi
#2}}
\providecommand{\BIBdecl}{\relax}
\BIBdecl

\bibitem{correll2018analysis}
N.~Correll, K.~E. Bekris, D.~Berenson, O.~Brock, A.~Causo, K.~Hauser, K.~Okada,
  A.~Rodriguez, J.~M. Romano, and P.~R. Wurman, ``Analysis and observations
  from the first amazon picking challenge,'' \emph{IEEE Transactions on
  Automation Science and Engineering}, vol.~15, no.~1, pp. 172--188, 2018.

\bibitem{eppner2016lessons}
C.~Eppner, S.~H{\"o}fer, R.~Jonschkowski, R.~Mart{\'\i}n-Mart{\'\i}n,
  A.~Sieverling, V.~Wall, and O.~Brock, ``Lessons from the amazon picking
  challenge: Four aspects of building robotic systems.'' in \emph{Robotics:
  Science and Systems}, 2016.

\bibitem{corbato2018integrating}
C.~H. Corbato, M.~Bharatheesha, J.~van Egmond, J.~Ju, and M.~Wisse,
  ``Integrating different levels of automation: Lessons from winning the amazon
  robotics challenge 2016,'' \emph{IEEE Transactions on Industrial
  Informatics}, 2018.

\bibitem{bicchi2000robotic}
A.~Bicchi and V.~Kumar, ``Robotic grasping and contact: A review,'' in
  \emph{ICRA}, vol. 348.\hskip 1em plus 0.5em minus 0.4em\relax Citeseer, 2000,
  p. 353.

\bibitem{shimoga1996robot}
K.~B. Shimoga, ``Robot grasp synthesis algorithms: A survey,'' \emph{The
  International Journal of Robotics Research}, vol.~15, no.~3, pp. 230--266,
  1996.

\bibitem{sahbani2012overview}
A.~Sahbani, S.~El-Khoury, and P.~Bidaud, ``An overview of 3d object grasp
  synthesis algorithms,'' \emph{Robotics and Autonomous Systems}, vol.~60,
  no.~3, pp. 326--336, 2012.

\bibitem{bohg2014data}
J.~Bohg, A.~Morales, T.~Asfour, and D.~Kragic, ``Data-driven grasp
  synthesis—a survey,'' \emph{IEEE Transactions on Robotics}, vol.~30, no.~2,
  pp. 289--309, 2014.

\bibitem{caldera2018review}
S.~Caldera, A.~Rassau, and D.~Chai, ``Review of deep learning methods in
  robotic grasp detection,'' \emph{Multimodal Technologies and Interaction},
  vol.~2, no.~3, p.~57, 2018.

\bibitem{saxena2007robotic}
A.~Saxena, J.~Driemeyer, J.~Kearns, and A.~Y. Ng, ``Robotic grasping of novel
  objects,'' in \emph{Advances in neural information processing systems}, 2007,
  pp. 1209--1216.

\bibitem{saxena2008robotic}
A.~Saxena, J.~Driemeyer, and A.~Y. Ng, ``Robotic grasping of novel objects
  using vision,'' \emph{The International Journal of Robotics Research},
  vol.~27, no.~2, pp. 157--173, 2008.

\bibitem{miller2003automatic}
A.~T. Miller, S.~Knoop, H.~I. Christensen, and P.~K. Allen, ``Automatic grasp
  planning using shape primitives,'' in \emph{Robotics and Automation, 2003.
  Proceedings. ICRA'03. IEEE International Conference on}, vol.~2.\hskip 1em
  plus 0.5em minus 0.4em\relax IEEE, 2003, pp. 1824--1829.

\bibitem{gualtieri2016high}
M.~Gualtieri, A.~ten Pas, K.~Saenko, and R.~Platt, ``High precision grasp pose
  detection in dense clutter,'' in \emph{Intelligent Robots and Systems (IROS),
  2016 IEEE/RSJ International Conference on}.\hskip 1em plus 0.5em minus
  0.4em\relax IEEE, 2016, pp. 598--605.

\bibitem{Pas2017Grasp}
A.~T. Pas, M.~Gualtieri, K.~Saenko, and R.~Platt, ``Grasp pose detection in
  point clouds,'' \emph{International Journal of Robotics Research}, vol.~36,
  no.~13, p. 027836491773559, 2017.

\bibitem{lenz2015deep}
I.~Lenz, H.~Lee, and A.~Saxena, ``Deep learning for detecting robotic grasps,''
  \emph{The International Journal of Robotics Research}, vol.~34, no. 4-5, pp.
  705--724, 2015.

\bibitem{zeng2017multi}
A.~Zeng, K.-T. Yu, S.~Song, D.~Suo, E.~Walker, A.~Rodriguez, and J.~Xiao,
  ``Multi-view self-supervised deep learning for 6d pose estimation in the
  amazon picking challenge,'' in \emph{Robotics and Automation (ICRA), 2017
  IEEE International Conference on}.\hskip 1em plus 0.5em minus 0.4em\relax
  IEEE, 2017, pp. 1386--1383.

\bibitem{zeng2018robotic}
A.~Zeng, S.~Song, K.-T. Yu, E.~Donlon, F.~R. Hogan, M.~Bauza, D.~Ma, O.~Taylor,
  M.~Liu, E.~Romo \emph{et~al.}, ``Robotic pick-and-place of novel objects in
  clutter with multi-affordance grasping and cross-domain image matching,'' in
  \emph{2018 IEEE International Conference on Robotics and Automation
  (ICRA)}.\hskip 1em plus 0.5em minus 0.4em\relax IEEE, 2018, pp. 1--8.

\bibitem{inproceedings}
A.~Causo, Z.-H. Chong, R.~Luxman, Y.~Yik~Kok, Z.~Yi, W.~C. Pang, R.~Meixuan,
  Y.~Seng~Teoh, W.~Jing, H.~Suratno~Tju, and I.-M. Chen, ``A robust robot
  design for item picking,'' 05 2018, pp. 7421--7426.

\bibitem{wade2017design}
S.~Wade-McCue, N.~Kelly-Boxall, M.~McTaggart, D.~Morrison, A.~W. Tow,
  J.~Erskine, R.~Grinover, A.~Gurman, T.~Hunn, D.~Lee \emph{et~al.}, ``Design
  of a multi-modal end-effector and grasping system: How integrated design
  helped win the amazon robotics challenge,'' \emph{arXiv preprint
  arXiv:1710.01439}, 2017.

\bibitem{redmon2016you}
J.~Redmon, S.~Divvala, R.~Girshick, and A.~Farhadi, ``You only look once:
  Unified, real-time object detection,'' in \emph{Proceedings of the IEEE
  conference on computer vision and pattern recognition}, 2016, pp. 779--788.

\bibitem{redmon2017yolo9000}
J.~Redmon and A.~Farhadi, ``Yolo9000: better, faster, stronger,'' \emph{arXiv
  preprint}, 2017.

\bibitem{ren2015faster}
S.~Ren, K.~He, R.~Girshick, and J.~Sun, ``Faster r-cnn: Towards real-time
  object detection with region proposal networks,'' in \emph{Advances in neural
  information processing systems}, 2015, pp. 91--99.

\bibitem{mahler2017dex}
J.~Mahler, J.~Liang, S.~Niyaz, M.~Laskey, R.~Doan, X.~Liu, J.~A. Ojea, and
  K.~Goldberg, ``Dex-net 2.0: Deep learning to plan robust grasps with
  synthetic point clouds and analytic grasp metrics,'' 2017.

\bibitem{mahler2017suction}
J.~Mahler, M.~Matl, X.~Liu, A.~Li, D.~Gealy, and K.~Goldberg, ``Dex-net 3.0:
  Computing robust robot suction grasp targets in point clouds using a new
  analytic model and deep learning,'' \emph{arXiv preprint arXiv:1709.06670},
  2017.

\bibitem{morrison2018cartman}
D.~Morrison, A.~W. Tow, M.~McTaggart, R.~Smith, N.~Kelly-Boxall, S.~Wade-McCue,
  J.~Erskine, R.~Grinover, A.~Gurman, T.~Hunn \emph{et~al.}, ``Cartman: The
  low-cost cartesian manipulator that won the amazon robotics challenge,'' in
  \emph{2018 IEEE International Conference on Robotics and Automation
  (ICRA)}.\hskip 1em plus 0.5em minus 0.4em\relax IEEE, 2018, pp. 7757--7764.

\bibitem{yu2016summary}
K.-T. Yu, N.~Fazeli, N.~Chavan-Dafle, O.~Taylor, E.~Donlon, G.~D. Lankenau, and
  A.~Rodriguez, ``A summary of team mit's approach to the amazon picking
  challenge 2015,'' \emph{arXiv preprint arXiv:1604.03639}, 2016.

\end{thebibliography}
\bibliographystyle{IEEEtran}

\end{document}